\documentclass[letterpaper, 10 pt, conference]{ieeeconf}  
\usepackage{amsfonts}
\usepackage{graphicx}
\usepackage[utf8]{inputenc}

\usepackage{booktabs}    
\usepackage{tabularx}    
\usepackage{ragged2e}    
\usepackage{multirow}    
\usepackage{array}    
\usepackage{caption}
\usepackage{float} 

\usepackage{setspace}

\IEEEoverridecommandlockouts                              

\usepackage{amsmath} 

\title{\LARGE \bf
CMMR-VLN: Vision-and-Language Navigation via Continual Multimodal Memory Retrieval
}

\author{Haozhou Li$^{1}$$^{,}$$^{3}$$^{*}$, Xiangyu Dong$^{1}$$^{*}$, Huiyan Jiang$^{3}$, Yaoming Zhou$^{4}$, Xiaoguang Ma$^{1}$$^{,}$$^{2}$$^{\dag}$
\thanks{$^{1}$The authors are with the Foshan Graduate School of Innovation at Northeastern University, Foshan, China}%
\thanks{$^{2}$The authors are with the Faculty of Robot Science and Engineering at Northeastern University, Shenyang, China}
\thanks{$^{3}$The authors are with the College of Software at Northeastern University, Shenyang, China}
\thanks{$^{4}$The authors are with the School of Aeronautic Science and Engineering at Beihang University, Beijing, China}%
\thanks{$^{*}$These authors contributed equally to this work.}%
\thanks{$^{\dag}$Corresponding author: Xiaoguang Ma {\tt\small (maxg@mail.neu.edu.cn)}}%
}

\begin{document}

\maketitle
\thispagestyle{empty}
\pagestyle{empty}


\begin{abstract}

Although large language models (LLMs) are introduced into vision-and-language navigation (VLN) to improve instruction comprehension and generalization, existing LLM-based VLN lacks the ability to selectively recall and use relevant priori experiences to help navigation tasks, limiting their performance in long-horizon and unfamiliar scenarios. In this work, we propose CMMR-VLN (Continual Multimodal Memory Retrieval based VLN), a VLN framework that endows LLM agents with structured memory and reflection capabilities. Specifically, the CMMR-VLN constructs a multimodal experience memory indexed by panoramic visual images and salient landmarks to retrieve relevant experiences during navigation, introduces a retrieved-augmented generation pipeline to mimick how experienced human navigators leverage priori knowledge, and incorporates a reflection-based memory update strategy that selectively stores complete successful paths and the key initial mistake in failure cases. Comprehensive tests illustrate average success rate improvements of 52.9\%, 20.9\% and 20.9\%, and 200\%, 50\% and 50\% over the NavGPT, the MapGPT, and the DiscussNav in simulation and real tests, respectively elucidating the great potential of the CMMR-VLN as a backbone VLN framework.

\end{abstract}


\section{INTRODUCTION}

Vision-and-language navigation (VLN) integrates computer vision and natural language processing to facilitate autonomous navigation of an agent \cite{anderson2018vision}. VLN necessitates the agent's comprehension of both the semantic content of instructions and visual data, thereby establishing a cross-modal mapping from language to spatial navigation. Studies on VLN not only drive advancements in multimodal intelligence but also offer fundamental underpinnings for applications like autonomous robot services and augmented reality interactions \cite{11128004}. 

Early research in VLN predominantly utilizes learning-based methodologies, training agents to seamlessly connect visual inputs and linguistic directives with actions \cite{fried2018speaker} \cite{wang2019reinforced}. While successful in familiar settings, these techniques rely heavily on annotated paths for guidance and frequently struggle to adapt to unfamiliar environments. Additionally, they lack capacity for unconstrained reasoning and integration of external information, resulting in limited adaptability in intricate real-world navigation contexts. 

Recent VLN studies introduce large language models (LLMs) \cite{hadi2023survey} to leverage their strong abilities to enhance instruction understanding and improve generalization in unfamiliar environments \cite{zhou2024navgpt} \cite{chen2024mapgpt}. However, unlike experienced human navigators who can instinctively recall relevant priori experiences in similar scenarios, deliberately use them to avoid suboptimal paths, and gradually become experts in certain navigation routes through accumulated experience, LLM-based VLN often lack the ability to identify and apply  relevant priori knowledge during navigation. Although LLMs possess vast general knowledge, they struggle to filter and ground this knowledge effectively in specific spatial contexts. Furthermore, their reasoning over navigation-relevant information often lacks structured logic \cite{long2024discuss}. This limits their ability to make coherent and context-aware decisions across long trajectories. 

To tackle these issues of priori LLM-based VLN approaches, we propose CMMR-VLN, a continual multimodal memory retrieval enhanced VLN, that endows LLM-based agents with retrieval and memory-based reasoning. Before navigation, the CMMR-VLN builds a viewpoint-level multimodal experience memory that stores and organizes past navigation cases for efficient retrieval. When facing a fork with visually similar candidate viewpoints, a standard LLM agent often struggles to choose correctly and may pick randomly, causing severe deviations. In contrast, an agent equipped with priori multimodal memory can retrieves the most relevant case to identify the viewpoint best aligned with the instruction. Beyond retrieval, the CMMR-VLN incorporates a reflection mechanism that evaluates each episode and selectively updates the experience memory, reinforcing successful routes in their entirety and distilling failures into concise notes tied to the first erroneous step, thereby enabling continual refinement and more consistent and context-aware decisions over time. 
\begin{figure*}[tbp]
    \centering
    \includegraphics[width=\textwidth]{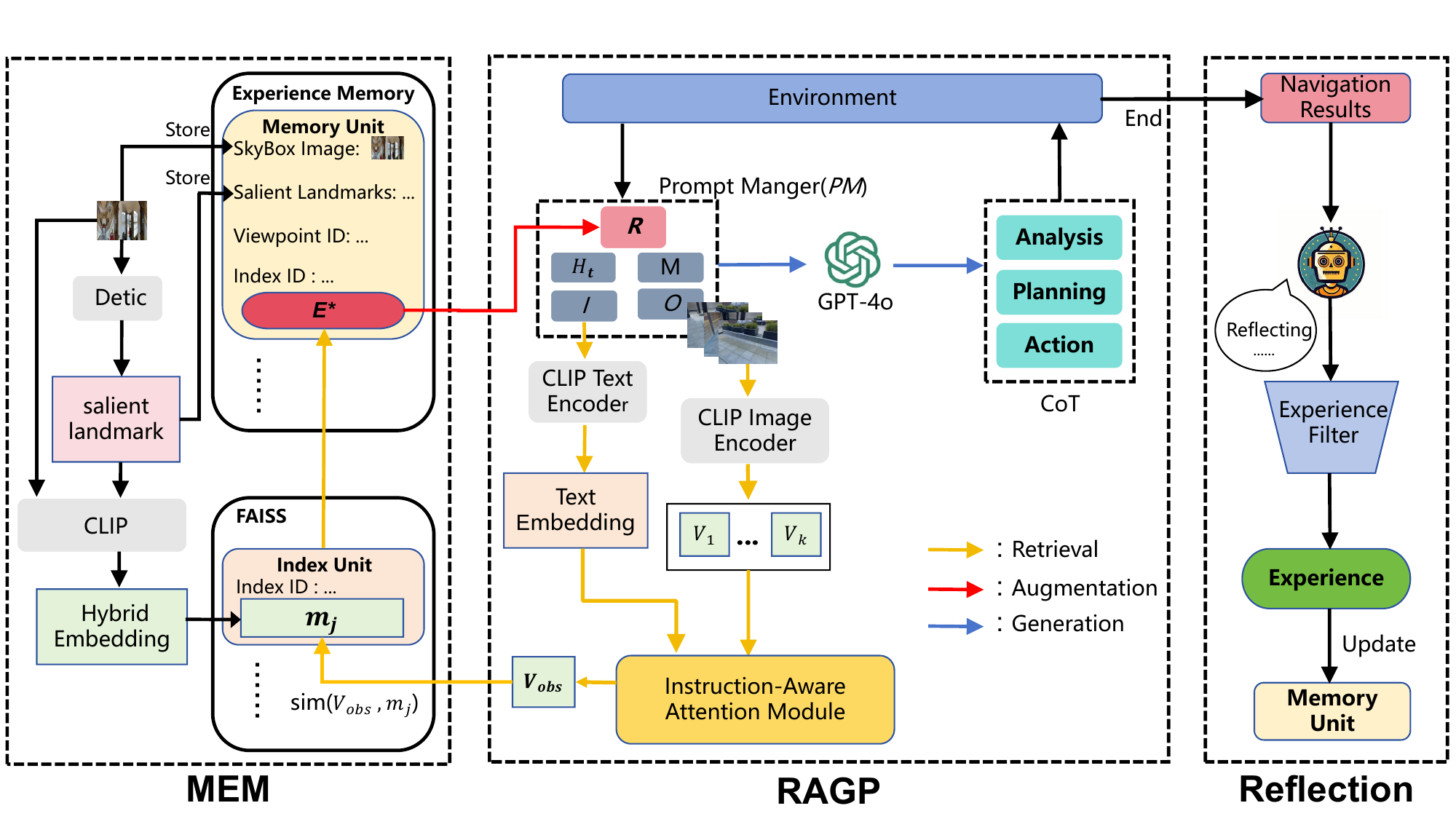}
    \caption{The overall CMMR-VLN framework consists of three modules from left to right. The Multimodal Experience Memory (MEM) performs memory building before navigation. The Retrieval-Augmented Generation Pipeline (RAGP) carries out corresponding prompting and action execution at each navigation step. The Reflection Module conducts reflection and updates the experience in memory after navigation.}
    \label{fig:your_figure}
\end{figure*}

Our contributions can be summarized as follows: 

\begin{itemize}
    \item We construct a structured multimodal experience memory, and enable retrieval-augmented reasoning that uses retrieved experiences as guiding rules for grounded and  step-by-step navigation.
    \item We design a reflection-based update module that reinforces successful trajectories and distills failure ones into key initial errors, enabling continual learning and efficient experience reuse.
    \item Comprehensive experiments show that the CMMR-VLN outperforms priori LLM-based SOTA VLN methods on the R2R dataset and in real-world tests, illustrating its great potential for various navigator applications.
\end{itemize}


\section{RELATED WORK}

\subsection{Vision-and-Language Navigation (VLN)}

Early VLN approaches predominantly rely on learning-based methods. The R2R \cite{anderson2018vision} introduces the task of following natural language instructions in photo-realistic environments with most initial methods adopting sequence-to-sequence architectures to map instructions to actions \cite{mei2016listen}. Subsequent VLN improvements incorporate cross-modal attention and panoramic representations \cite{fried2018speaker}, as well as imitation learning techniques \cite{wang2019reinforced}, to better align visual and linguistic modalities to improve navigation accuracy. 

Meanwhile, pre-trained vision-language models (VLMs) such as BERT \cite{devlin2019bert} and CLIP \cite{radford2021learning} are introduced to enhance cross-modal grounding. For example, PREVALENT \cite{hao2020towards} employs pre-training strategies on large-scale navigation data to improve instruction grounding. VLN-BERT \cite{hong2021vln} and REVERIE \cite{qi2020reverie}  leverage pre-trained vision-language transformers for more effective visual grounding and object-level reasoning.

More recently, large language models (LLMs) are incorporated into VLN to leverage their powerful reasoning and generalization abilities. The NavGPT \cite{zhou2024navgpt} formulates VLN as a step-wise instruction-following task using GPT-3 with historical context and candidate views for long-horizon planning. The DiscussNav \cite{long2024discuss} introduces deliberative reasoning by prompting the agent to “discuss” each action step with a LLM, enabling more reflective decision-making. The MapGPT \cite{chen2024mapgpt} uses frontier semantic maps to enhance spatial reasoning. While these LLM-based methods demonstrate promising results, they lack the capabilities to fully leverage past multimodal experiences.

\subsection{Memory-Augmented and Reflective Agents }

To enhance reasoning in complex multimodal VLN tasks, a growing body of research explores memory-based or retrieval-augmented frameworks. Retrieval-augmented generation (RAG)  and its extensions combine generation with external knowledge retrieval to provide context-relevant information at each decoding step in both text-only and multimodal tasks \cite{lewis2020retrieval,guu2020retrieval}.

In the VLMs field, multimodal RAG methods have been developed to enable retrieval of image-text pairs from large-scale databases, supporting grounded question answering and visual planning. Chain-of-Thought \cite{wei2022chain} prompting further improves reasoning by encouraging step-by-step thought processes, which can be combined with retrieval to support complex decision-making \cite{kojima2022large}. Moreover, reflection mechanisms  allow agents to assess priori decisions and revise plans based on environmental feedback \cite{shinn2023reflexion}. In navigation contexts, this can enable LLM agents to update goals or instructions dynamically based on changes in spatial context or agent performance \cite{yao2023react}.


    
    

\section{METHODOLOGY}
Inspired by human navigation behaviors, we propose the CMMR-VLN, a zero-shot VLN framework based on continual multimodal memory retrieval and experience reflection. The overall framework of the CMMR-VLN is
illustrated in Fig 1. In this section, we describe how we construct the multimodal experience memory before navigation, how the agent leverages retrieved experiences at each navigation step, and introduce the post-navigation reflection mechanism that updates the memory for continual improvement. 

\subsection{Multimodal Experience Memory (MEM)} 
When humans navigate in complex unfamiliar environments , they often rely on salient landmarks and recall familiar routes \cite{chen2019learning,lynch1960image} to orient themselves. Inspired by this, we construct a MEM module that supports accurate retrieval of relevant past experiences based on current scenes \cite{khandelwal2019generalization}. 
Specifically, we construct an experience memory organized as a set of memory units, each corresponding to a unique viewpoint in the Matterport3D simulator \cite{chang2017matterport3d}. For every viewpoint, we store the panoramic SkyBox image and its associated viewpoint ID within the memory units. To enrich the memory with semantic cues, we apply a fine-tuned Detic model \cite{zhou2022detecting} to detect salient landmarks from the panoramic images and store the salient landmark texts in the same units as shown in Fig 1.

To enable efficient and accurate retrieval \cite{majumdar2020improving} , we encode both the panoramic images and their associated salient landmark texts using a CLIP model \cite{radford2021learning}, generating a hybrid image-text embedding for each viewpoint. These embeddings are indexed using FAISS \cite{johnson2019billion} as shown in Fig 1. A unique index ID is used to link each index unit to its corresponding memory unit, allowing bidirectional lookup between the embedding space and the stored experience memory. The detailed content of the experience within each memory unit will be introduced in Section C.


\subsection{Retrieval-Augmented Generation Pipeline (RAGP) }
To implement continual multimodal memory retrieval, we design a  RAGP at each navigation step \cite{yao2023react} as shown in Fig 1. 
At the beginning of each navigation step, the navigation instruction \textit{I} is inserted into the prompt manger as the high-level goal, then the prompt manger receives RGB observations \textit{O} $\{ o_1, o_2, \ldots, o_K \}$ for all \textit{K} candidate viewpoints accessible from the current location. These images are provided by the Matterport3D Simulator \cite{chang2017matterport3d}. To maintain temporal continuity and reasoning consistency, we append the historical trajectory context $ H_t $  to the prompt manger. Here, $ H_t $ contains a sequence of previously visited viewpoint IDs and the reasoning behind each choice:

\begin{equation}
    H_{t}=\left\{\left(\mathrm{id}_{1}, r_{1}\right),\left(\mathrm{id}_{2}, r_{2}\right), \ldots,\left(\mathrm{id}_{t-1}, r_{t-1}\right)\right\}
\end{equation}
where $ r_i $ is the textual rationale for selecting viewpoint $ id_i $ .

Following the methodology in the MapGPT \cite{chen2024mapgpt}, we construct a semantic topological map and insert it into the prompt manger:
\begin{equation}
    M=(V, E)
\end{equation}

Where each node $v \in V$  corresponds to a viewpoint enriched with semantic features, and each edge  $e \in E$ denotes a navigable connection between viewpoints. 

Rather than building the entire graph upfront, the map is expanded incrementally during navigation. At each step, the current viewpoint’s navigable neighbors are identified, their semantic features are extracted, and new nodes and edges are appended to \textit{M} . This dynamic map allows the LLM to maintain an up-to-date representation of spatial connectivity and to leverage global route structures for multi-step planning, rather than relying solely on local observations.


As shown in Fig 1, we encode the instruction and the set of candidate viewpoint images from the prompt manger using the CLIP \cite{radford2021learning} text and image encoders, respectively. The resulting embeddings are fused by an instruction-aware attention module to produce a single observation embedding $v_{\text{obs}}$ that integrates multi-view context while focusing on viewpoints most relevant to the navigation goal: 
\begin{equation}
    \alpha_{k}=\operatorname{softmax}\left(u^{\top} W v_{k}\right), \quad v_{\mathrm{obs}}=\sum_{k=1}^{K} \alpha_{k} v_{k}
\end{equation}
where $u$ is the embedding of the navigation instruction, and $W$ is a learned projection matrix \cite{vaswani2017attention}.

Compared to naive mean pooling, this instruction-aware attention handles cases more effectively where some candidate viewpoints lie in peripheral or less informative directions from the current location \cite{bahdanau2014neural}. Mean pooling can dilute the influence of relevant viewpoints, potentially reducing retrieval accuracy. In contrast, the attention mechanism assigns higher weights to candidate viewpoints that are semantically aligned with the instruction, producing a fused observation embedding $v_{\text{obs}}$ that is more discriminative for retrieving the most relevant past experiences.

We then compute cosine similarity between $v_{\text{obs}}$ and the multimodal hybrid embeddings $\{m_j\}$ stored in the FAISS index units \cite{johnson2019billion}:
\begin{equation}
    \operatorname{sim}\left(v_{\mathrm{obs}}, m_{j}\right)=\frac{v_{\mathrm{obs}} \cdot m_{j}}{\left\|v_{\mathrm{obs}}\right\|\left\|m_{j}\right\|}
\end{equation}
wherein the most relevant priori experience of memory unit \cite{borgeaud2022improving}.

\( E^* \) is transformed into an explicit
\begin{figure*}[htbp!]
    \centering
    \includegraphics[width=\textwidth]{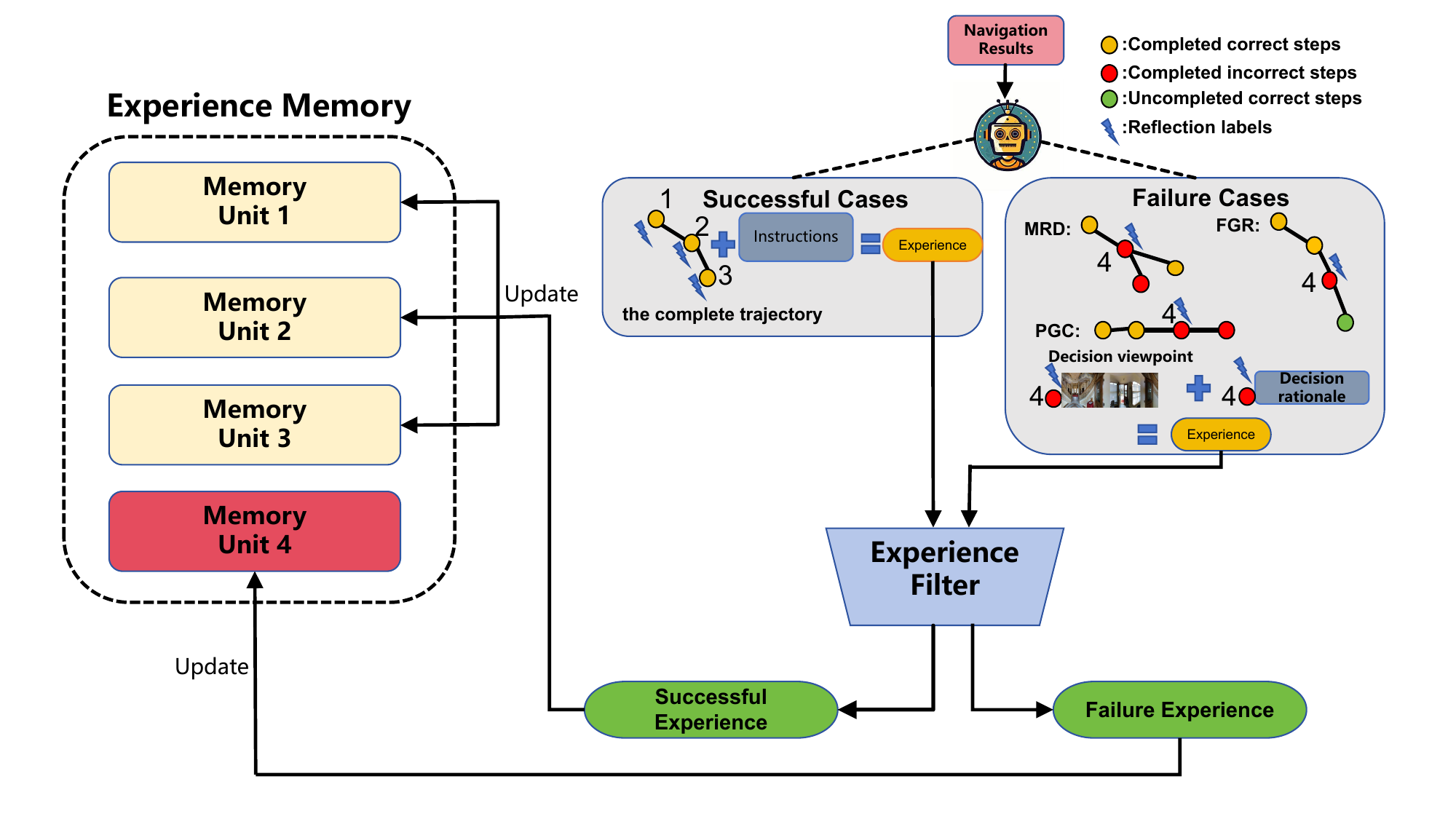}
    \caption{Details of the Reflection Module in Fig 1.}
    \label{fig:your_figure}
\end{figure*}
navigation rule R that incorporates priori experience relevant to the current context. This rule R is then augmented into the prompt manager as a high-priority constraint, ensuring the LLM prioritizes it over other contextual information during decision-making \cite{lewis2020retrieval}.

As shown in Fig 1, the prompt manger can be represented as:
\begin{equation}
    \mathcal{PM}=\{I, O, H, M ; R\}
\end{equation}
where \( I \) is the navigation instruction, \( O \) is the images of candidate viewpoints, \( H \) is the navigation history, and \( M \) is the semantic topological map. The semicolon here emphasizes that \( R \) is not merely another contextual element but a guiding principle that shapes the entire reasoning process.
This prompt manager \( PM \) serves as the input to the LLM, which generates a structured chain-of-thought output consisting of analysis, planning, and action \cite{wei2022chain}:

\begin{itemize}
\item Analysis: Thoughts on the current navigation state while explicitly integrating the guidance from \( R \).

\item Planning: A multi-step plan
\begin{equation}
    P_{t}=\left\{\left(a_{1}, g_{1}\right),\left(a_{2}, g_{2}\right), \ldots\right\}
\end{equation}
where each action \( a_i \) is paired with its intended goal \( g_i \).

\item Action: Selection of the next viewpoint \( id_t \) for execution.

\end{itemize}

By assigning \( R \) a privileged role in the prompt manger, the reasoning process is consistently steered by priori navigation experience in similar contexts, leading to more reliable and contextually aligned decision making \cite{shinn2023reflexion}.

\subsection{Reflection and Memory Update }

At the end of each navigation episode, a reflection is activated to evaluate the results and update the experience memory accordingly. The evaluation first determines whether the final position lies within the vicinity of the navigation target, classifying the episode as a success or failure \cite{mnih2015human}. 

To minimize irrelevant noise in the experience base and avoid overwhelming the LLM with unnecessary information, different update strategies are applied based on the classification \cite{anderson1982acquisition} as illustrated in Fig 2: 

\begin{itemize}
    \item \textbf{Successful cases}:

 The navigation instruction and the complete trajectory are inserted into the experience memory unit corresponding to every viewpoint along the path. This strategy reflects the way humans tend to remember an entire successful route in similar situations, rather than only focusing on a few key points.
    \item \textbf{Failure cases}:

 We focus on the first incorrect decision step, which is identified as one of three failure types:
    \begin{enumerate}
        \item \textbf{Mid-route deviation (MRD)} – deviating from the correct path before reaching the goal.
        \item \textbf{False goal recognition (FGR)} – incorrectly assuming arrival at the destination.
        \item \textbf{Post-goal continuation (PGC)} – continuing to move past the actual destination.
    \end{enumerate}
\end{itemize}

For each case, the corresponding decision viewpoint and decision rationale are extracted from the historical context $ H $ and stored in the experience memory unit of that viewpoint. Additional textual annotations are appended to indicate the specific error type. To emulate the human tendency to vividly remember the first wrong step in a failure route, the panoramic image of that viewpoint is also included in the retrieved rule \textit{R} if this failure experience is later accessed during navigation.

As illustrated in Fig 2, we apply an experience filter to interact with the Experience Memory. Specifically, for successful experiences, if the memory already contains a similar successful route with better efficiency (shorter or more optimal), the new one is discarded. Otherwise, it replaces the less optimal version.
For failure experiences, if both the decision point and reason are already represented in the memory, the new entry is ignored.

\section{EXPERIMENTS AND ANALYSIS}
\subsection{Experiment Setup}
We employ GPT-4o \cite{hurst2024gpt} as the backbone LLM in the CMMR-VLN, leveraging its strong capability in multimodal perception and reasoning. Experiments are conducted in the Matterport3D simulator \cite{chang2017matterport3d} with the Room-to-Room (R2R) dataset \cite{anderson2018vision}, where each trajectory pairs a natural language instruction with a start and goal location in photo-realistic indoor environments.

We evaluate performance using four standard VLN metrics: Navigation Error (NE) is the average distance from the final position to the goal; Success Rate (SR) is the proportion of episodes ending within 3m of the target, Oracle Success Rate (OSR) counts success if any visited viewpoint is within 3m of the goal, and Success weighted by Path Length (SPL) balances success with trajectory efficiency.

 \begin{table}[h]
    \centering
    \caption{Results on R2R Validation Unseen Split.}
    \label{tab:zero-shot_comparison}
    \resizebox{\columnwidth}{!}{ 
    \begin{tabular}{l l c c c c}
        \hline
        Setting & Method & NE↓ & OSR↑ & SR↑ & SPL↑ \\
        \hline
        \multirow{4}{*}{
        Training-Free
        } & NavGPT \cite{zhou2024navgpt}        & 6.46 & 42 & 34 & 29 \\
                                   & MapGPT \cite{chen2024mapgpt}        & 5.63 & 57 & 43 & 34 \\
                                   & DiscussNav \cite{long2024discuss}    & 5.32 & 61 & 43 & 40 \\
                                   & CMMR-VLN(Ours)  & \textbf{5.10} & \textbf{63} & \textbf{52} & \textbf{51} \\
        \hline
    \end{tabular}
    }
\end{table}

\subsection{Simulation Experiments}

We conduct quantitative experiments on the R2R validation unseen split, which contains 11 scenes and 783 trajectories. 
As shown in Table $\mathrm{I}$, the CMMR-VLN obtains a SR improvement of 52.9\% over the NavGPT, demonstrating the superior benefits of retrieval-augmented reasoning. The CMMR-VLN achieves a 50\% improvement in SPL over the MapGPT, attributing to its semantic topological map, which encourages broader global exploration and can lengthen trajectories. Meanwhile, the CMMR-VLN obtains improvements of 20.9\% and 27.5\% in SR and SPL over the DiscussNav respectively. Notably, the DiscussNav relies on multiple LLMs as specialized experts, whereas the CMMR-VLN employs a single LLM as a decision expert, reducing computational overhead while delivering superior performance.

\subsection{Ablation Study}
All experiments are performed on a sampled subset of the R2R dataset \cite{anderson2018vision} containing 72 environments and 216 trajectories.
We first investigate the role of retrieved experiences as explicit navigation rules. Instead of asking the LLM to reason about the relationship between retrieved experiences and other contextual inputs, we treat the retrieved experiences as ordinary contextual information, concatenated with the instruction, observation images, and other such information. As shown in Table \textnormal{II}, the CMMR-VLN without navigation rules performs significantly worse than the full CMMR-VLN across all metrics, particularly with a drop of over 10 SPL points \cite{lewis2020retrieval,wei2022chain}. Interestingly, its performance is almost identical to that of the MapGPT \cite{chen2024mapgpt}, suggesting that the LLM largely ignore them and fall back to standard contextual signals without explicit reasoning on how retrieved experiences guide decision-making.

In addition, we remove the reflection-based update mechanism \cite{shinn2023reflexion} and replace the stored experiences with fixed scene-level textual descriptions of each viewpoint. As shown in Table \textnormal{II}, the CMMR-VLN with Scene Description shows the lowest performance for all metrics. Qualitative inspection of the LLM’s reasoning outputs reveals that scene descriptions distract the agent. In fact, the LLM tends to overemphasize aligning the retrieved text with the current observations rather than focusing on navigation. This misalignment often leads to irrational behaviors such as circling in place, resulting in inefficient and less successful navigation.

\begin{table}[t]
    \centering
    \caption{Ablation Study Results.}
    \label{tab:example}
    \resizebox{0.85\columnwidth}{!}{  
        \begin{tabular}{lcccc}
            \toprule
            Method & NE↓ & OSR↑ & SR↑ & SPL↑ \\
            \midrule
            CMMR-VLN & \textbf{5.08} & \textbf{65} & \textbf{56} & \textbf{50.91} \\
            MapGPT \cite{chen2024mapgpt} & 5.62 & 57.9 & 47.7 & 38.1 \\
            CMMR-VLN w/o NR & 5.59 & 56.5 & 48.2 & 40 \\
            CMMR-VLN w/ SD & 10.20 & 40 & 40 & 37.63 \\
            \bottomrule
        \end{tabular}
    }
\end{table}

\begin{figure}[h]
    \centering
    \parbox{3.4in}{
        \centering 
        \includegraphics[width=\linewidth]{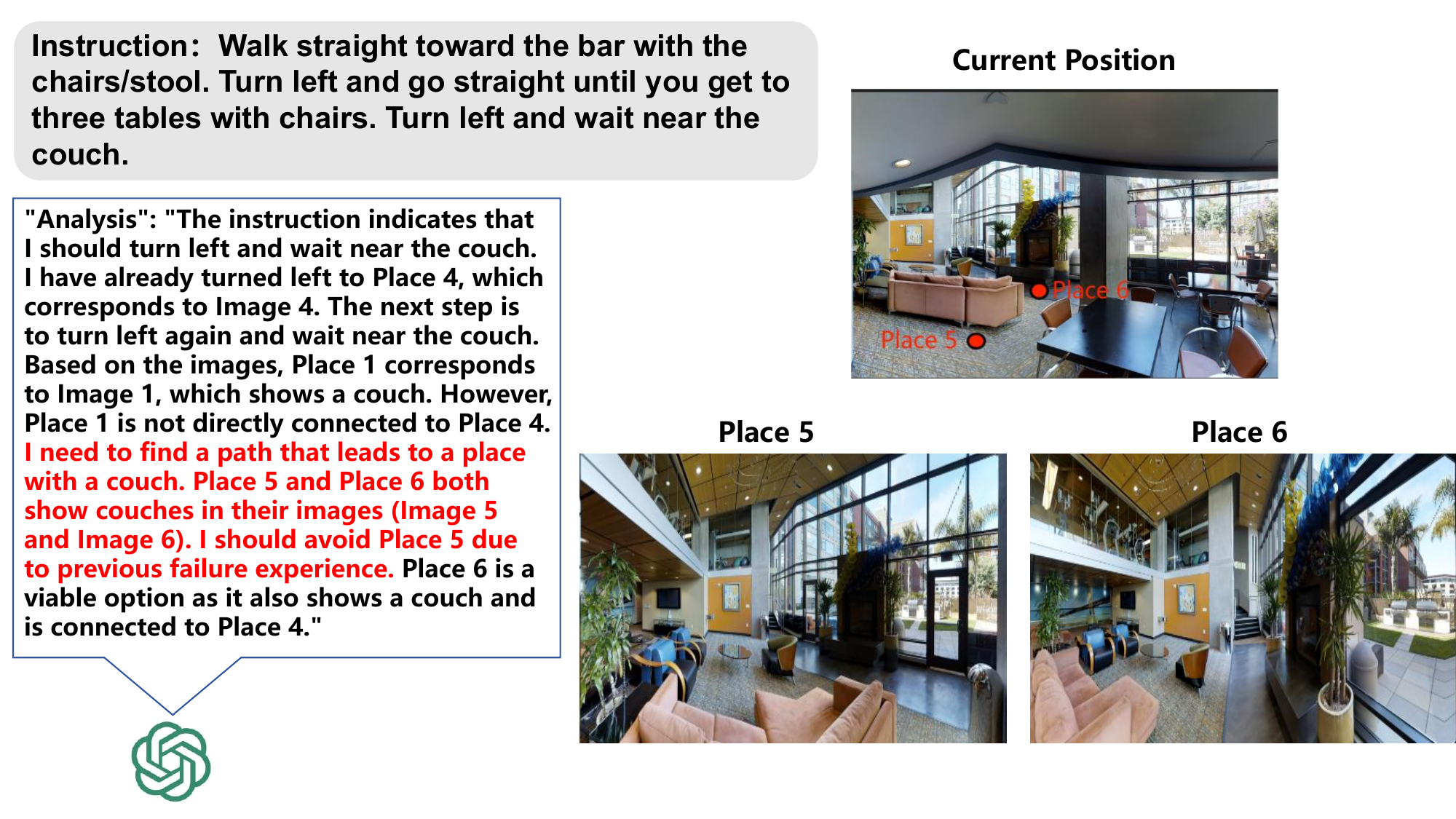} 
    }
    \caption{Case study 1.}
    \label{figurelabel}
\end{figure}

\subsection{Case Study}
In this section, we analyze two R2R navigation cases to illustrate the importance of leveraging past experiences in guiding current decision-making. The first case presents a scenario where the agent is located at the current position and needs to execute the instruction: \textit{“turn left again and wait near the couch.”} As illustrated in Fig 3, the LLM-based reasoning process identifies two potential candidates, Place 5 and Place 6, both of which contain couches that match the instruction. However, due to the limited spatial understanding of “turn left” in the LLM’s reasoning, the agent cannot rely solely on geometric interpretation to select the correct direction. At this point, past navigation experiences become helpful. Specifically, based on priori failure experience, the agent recalls that choosing Place 5 previously leads to unsuccessful navigation. Consequently, the reasoning process, highlighted in red in Fig 3, concludes that Place 6 is the preferable option, since it not only satisfies the semantic requirement of containing a couch but also avoids repeating past mistakes. This case demonstrates how incorporating past failures enables the agent to refine its decision-making and achieve more reliable navigation. 

The second case study is shown in Fig 4. The instruction requires the agent to stop at the doorway, while the doorway is not directly visible from the current positions. In this situation, the agent leverages priori successful experience of executing similar instructions in the same environment. Guided by this priori trajectory, GPT-4o infers that moving toward Place 12 is the most promising choice. This hypothesis is then validated using both the observation at Place 12 and the semantic topological map,  revealing a table followed by a doorway and confirming alignment with the instruction.
\begin{figure}[thpb]
    \centering
    \parbox{3.4in}{
        \centering 
        \includegraphics[width=\linewidth]{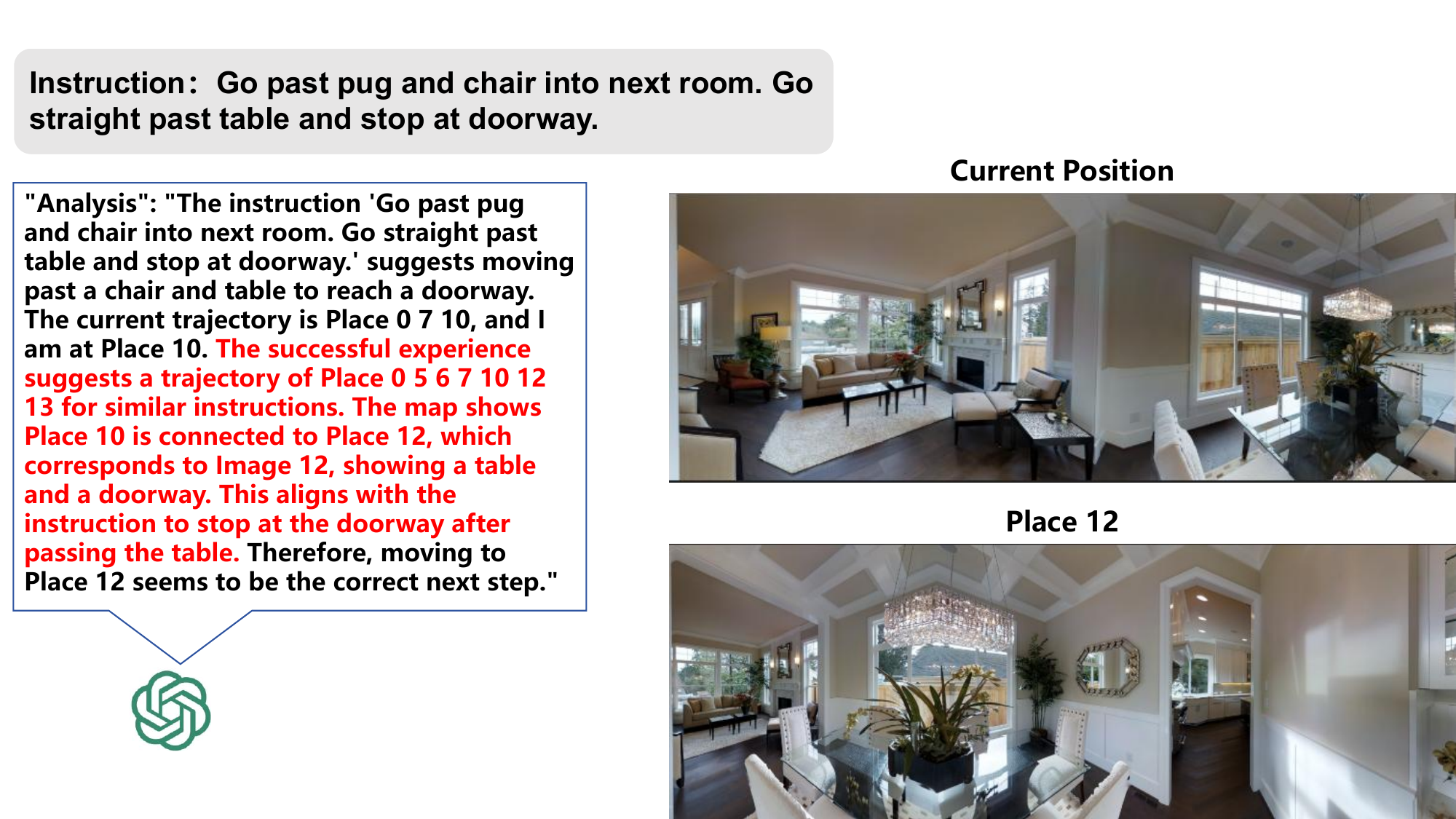} 
    }
    \caption{Case study 2.}
    \label{figurelabel}
\end{figure}

\subsection{Experiments on A Real Robot}
We further evaluate the CMMR-VLN in real-world scenarios using a TurtleBot 4 Lite mobile robot equipped with an RGB camera mounted 50 cm above the base. To adapt our method to continuous environments, we partition each indoor scene into several regions based on prominent landmarks or large objects, treating each region as a memory unit.
Unlike the discrete viewpoint actions in Matterport3D simulator \cite{chang2017matterport3d}, we define low-level motion primitives for real-world navigation, e.g., "Move forward 50 cm", "Turn left 45°". We design 20 natural language instructions across different environments, including REVERIE \cite{qi2020reverie} style instructions such as “First locate the trash can, then move to its side.”in Table III

Real robot tests on TurtleBot 4 Lite show that the CMMR-VLN obtains SR improvements of 200\%, 50\% and 50\% over the NavGPT, the MapGPT, and the DisscussNav, respectively. In fact, detailed experimental results show that the NavGPT can generally follow human instructions to navigate in real environments, but its single-round reasoning limits its ability to complete long-horizon or abstract tasks. The MapGPT benefits from semantic topological maps in simulated discrete environments, yet struggles to fully exploit them in real-world continuous environments where region connectivity is more complex. The DiscussNav makes decisions through multi-agent discussion, but it requires multiple steps to complete complex instructions, which significantly increases API costs while providing only marginal performance gains. In contrast, our CMMR-VLN consistently achieves superior performance, owing to its ability to continuously retrieve multimodal experiences updated via reflection during real-world navigation, enabling more effective decision-making in complex environments.

\begin{table}[h]
    \centering
    \caption{Navigation Task Comparsion on A Real Robot.}  
    \label{tab:zero-shot_comparison_sr}
    \resizebox{0.67\columnwidth}{!}{
        \begin{tabular}{l l c}  
            \hline
            Setting & Method & SR↑ \\  
            \hline
            \multirow{4}{*}{
        Training-Free
            } & NavGPT \cite{zhou2024navgpt}        & 10 \\
                                       & MapGPT \cite{chen2024mapgpt}        & 20 \\
                                       & DiscussNav \cite{long2024discuss}    & 20 \\
                                       & CMMR-VLN(Ours)  & \textbf{30} \\  
            \hline
        \end{tabular}
    }
\end{table}

\section{CONCLUSIONS}
In this work, we propose the CMMR-VLN, a retrieval-augmented  continual vision-and-language navigation framework that leverages priori experiences to guide decision-making. Retrieved experiences are transformed into explicit navigation rules, while a reflection mechanism continuously updates the experience memory with successful and failure cases. Experiments demonstrate significant improvements, particularly in the zero-shot setting, and case studies highlight how past experiences enable more accurate and efficient navigation. While the CMMR-VLN shows strong performance, its effectiveness still depends on the quality and scope of stored experiences in memory. Future work will focus on expanding the experience memory and improving retrieval efficiency to further enhance its adaptability and generalization.

\bibliographystyle{IEEEtran.bst}
\bibliography{conference}

\end{document}